\documentclass[lettersize,journal]{IEEEtran}
\usepackage{amsmath,amsfonts}
\usepackage{algorithm}
\usepackage{algpseudocode}   
\usepackage{booktabs}  
\usepackage{algorithm}
\usepackage{array}
\usepackage[caption=false,font=normalsize,labelfont=sf,textfont=sf]{subfig}
\usepackage{textcomp}
\usepackage{stfloats}
\usepackage{amssymb}

\usepackage{url}
\usepackage{verbatim}
\usepackage{graphicx}
\usepackage{cite}
\begin{document}

\title{UniSkip-Mamba: A Frequency-Aware State Space Model for Audio-Visual Temporal Forgery Localization}

\author{
    Cangjin Yu\IEEEauthorrefmark{1},
    Quan Zhang\IEEEauthorrefmark{1},
    Dan Jiang,
    and Ke Zhang\IEEEauthorrefmark{2}

\thanks{\IEEEauthorrefmark{1}Cangjin Yu and Quan Zhang contributed equally to this work.}%
\thanks{\IEEEauthorrefmark{2}Corresponding author: Ke Zhang.}%
\thanks{Cangjin Yu and Ke Zhang are with Soochow University, Suzhou, China.}%
\thanks{Quan Zhang and Dan Jiang are with Tsinghua University, Beijing, China.}%
}



\maketitle
\begin{abstract}
With the proliferation of AI-generated content, sophisticated 
multimedia manipulation has raised critical concerns about malicious 
applications such as opinion manipulation and evidence fabrication, 
making Audio-Visual Temporal Forgery Localization (AV-TFL) an urgent 
research frontier. Existing TFL methods have progressed along two 
main paradigms: Transformer-based temporal modeling and channel-wise 
multimodal fusion. While these approaches capture temporal dependencies 
and cross-modal correlations, they process all frequency components 
indiscriminately, leading to overfitting on high-frequency noise and 
limited robustness under real-world data degradation. Through systematic 
frequency domain analysis, we find that forgery-discriminative patterns 
concentrate in the low/mid-frequency range (normalized frequency 
0--0.15), while high-frequency components primarily introduce noise—removing 
them even improves detection performance by +1.4\%. Based on this 
phenomenon, we propose UniSkip-Mamba, a frequency-aware State Space 
Model framework that incorporates Unified Multimodal Sequence Fusion 
to preserve cross-modal phase relationships, and Skip-Scanning Mamba Blocks that implement frequency-aware regularization 
through a novel Group-Scan-Merge mechanism, naturally biasing learning toward discriminative low/mid-frequency patterns (0--0.15) while maintaining representational completeness. We achieve state-of-the-art (SOTA) performance: 63.4\% AP@0.95 on LAV-DF (+9.8\% improvement) and 63.58\% mAP on AV-Deepfake1M (+14.32\% improvement), with 6$\times$ faster inference. Our frequency-domain analysis provides theoretical justification from a signal processing perspective for why skip-scanning inherently improves both accuracy and robustness.
\end{abstract}

\begin{figure}[h!]
    \centering
    \includegraphics[width=0.5\textwidth]{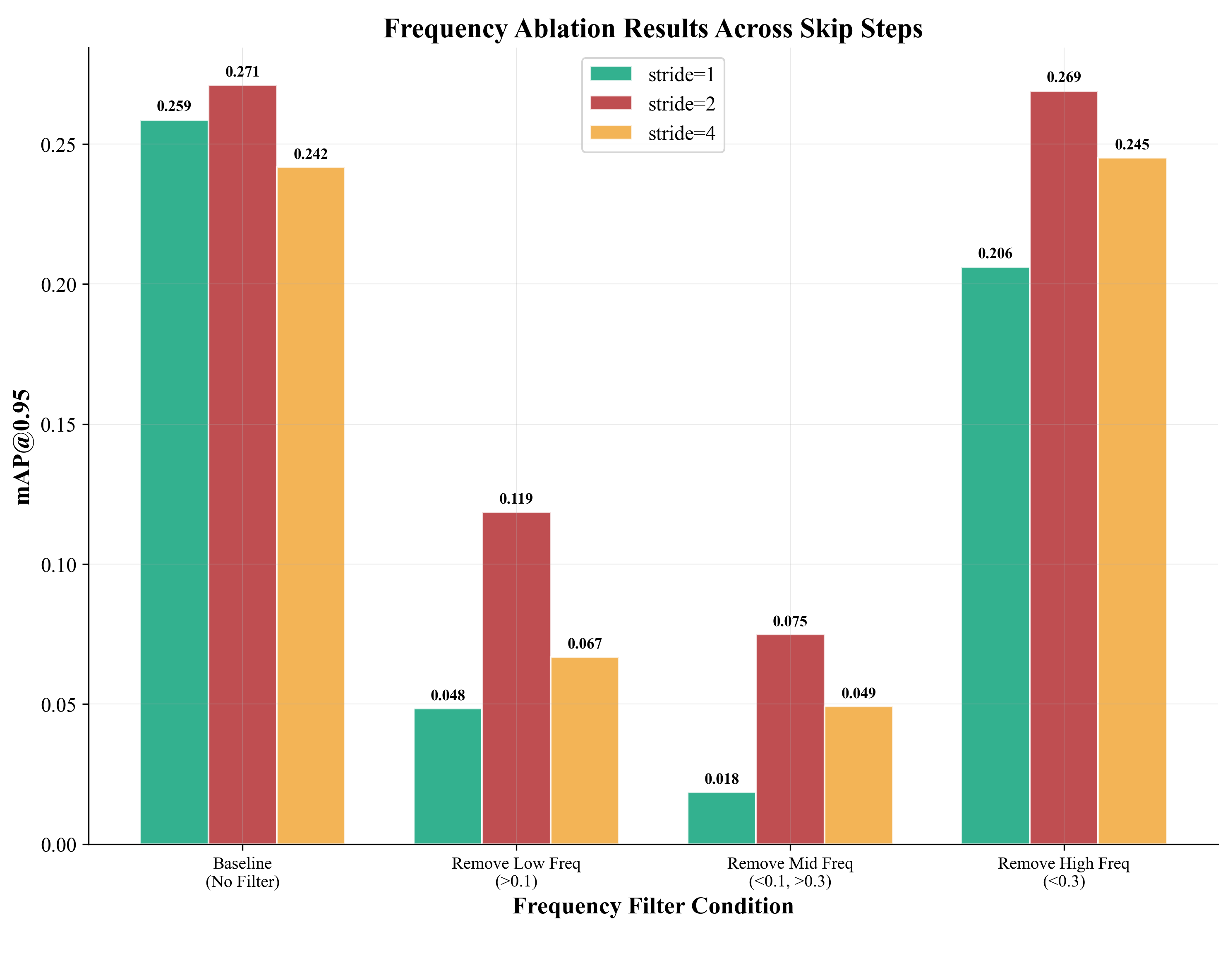}
    \caption{\textbf{Motivating Observation: Frequency Band Contribution Analysis.} We systematically remove each frequency band from input features and measure performance impact (mAP@0.95). \textbf{Key findings:} (1) Low/mid-frequency bands (0--0.15) are critical for detection, with removal causing up to 92.8\% performance drop; (2) High-frequency removal ($>0.15$) has minimal impact ($-0.7\%$ to $+1.4\%$), with stride-4 models actually improving—confirming that high frequencies introduce noise rather than discriminative information. This motivates our Skip-Scanning design as a frequency-aware regularization strategy. Comprehensive experimental setup and detailed analysis in Section~\ref{ssec:freq_analysis}.
}
    \label{fig:freq_contribution}
\end{figure}
\section{Introduction}

The proliferation of Artificial Intelligence Generated Content (AIGC)~\cite{aldausari2022video,cai2022devit,wang2022fastlts} has enabled sophisticated multimedia manipulation, raising critical concerns about malicious applications such as opinion manipulation and evidence fabrication. While the multimedia forensics community has made significant progress in Deepfake detection~\cite{deng2022detection,kwak2022low,zhang2022deepfake,zhao2021multi,zhou2021joint}, existing methods predominantly rely on binary classification, failing to identify where manipulation occurs temporally. This limitation severely constrains their utility in practical scenarios such as judicial forensics and content moderation, where precise temporal boundaries of forged segments are indispensable. Consequently, Audio-Visual Temporal Forgery Localization (AV-TFL) has emerged as a critical research frontier.

Before addressing the technical challenges of AV-TFL, we pose a fundamental question: \textit{In which frequency bands does discriminative forgery information reside?} Through spectral analysis and systematic ablation experiments, we reveal that forgery-discriminative patterns concentrate in low/mid-frequency bands (normalized frequency 0--0.15)
. Removing these bands causes 56--93\% performance drop, while removing high frequencies ($>$0.15) has negligible impact---stride-4 models even improve by +1.4\% when high frequencies are filtered. This demonstrates that high-frequency components, which capture frame-level variations such as compression artifacts, primarily introduce noise rather than discriminative information.

This frequency-domain insight exposes a critical blind spot in existing approaches: current methods process all frequency components indiscriminately, capturing both useful low-frequency patterns and harmful high-frequency noise. As we demonstrate in Section~\ref{ssec:robustness}, this leads to severe robustness degradation under real-world quality losses. Furthermore, existing TFL methods~\cite{zhang2023ummaformer,yin2025context} employ channel-wise concatenation that rigidly aligns modalities, preventing detection of asynchronous artifacts. Whether using Transformers~\cite{zhang2022actionformer} or recent Mamba adaptations~\cite{zhu2024vision,chen2024video}, they all employ dense scanning, amplifying the high-frequency noise identified above.

To address these challenges, we propose \textbf{UniSkip-Mamba}, a framework \textbf{grounded in frequency-principled design}. It leverages the \textbf{Mamba backbone} with $O(T)$ complexity to efficiently model the extended sequences where low-frequency patterns manifest. Central to our design is the \textbf{Unified Sequence Representation}, which serializes audio-visual features to preserve cross-modal phase relationships, enabling detection of asynchronous forgeries with arbitrary temporal offsets. Furthermore, we introduce \textbf{Skip-Scanning via Group-Scan-Merge}, a novel mechanism implementing structural low-pass regularization. By aligning the scanning stride with the discriminative band (0--0.15) identified above, it naturally filters high-frequency noise while retaining representational completeness, achieving simultaneous improvements in accuracy, speed ($2\times$), and robustness.

The main contributions of this paper are summarized as follows:
\begin{itemize}
    \item We propose \textbf{UniSkip-Mamba}, an efficient SSM framework achieving linear complexity $O(T)$, effectively overcoming Transformer bottlenecks for long-range temporal modeling in AV-TFL.

    \item We present \textbf{comprehensive frequency-principled analysis} combining theoretical predictions with empirical validation, revealing that discriminative information concentrates in low/mid-frequency bands (0--0.15) and theoretically justifying stride-2 as optimal structural regularization.

    \item We achieve \textbf{SOTA performance}: 63.4\% AP@0.95 on LAV-DF (+9.8\%) and 63.58\% mAP on AV-Deepfake1M (+14.32\%), with $6\times$ faster inference and superior robustness.
\end{itemize}
\section{related work}
\subsection{Temporal Forgery Localization}
Early multimedia forensics approaches~\cite{deng2022detection,kwak2022low,
zhang2022deepfake,zhao2021multi,zhou2021joint} focused primarily on 
binary classification, often failing to provide precise temporal 
boundaries. To address this, architectures from Temporal Action 
Localization (TAL), such as ActionFormer~\cite{zhang2022actionformer} 
and TriDet~\cite{shi2023tridet}, were repurposed as strong baselines. 
Subsequently, dedicated TFL frameworks emerged to tackle specific 
forgery challenges: UMMAFormer~\cite{zhang2023ummaformer} integrates 
temporal abnormality attention; DiMoDif~\cite{koutlis2024dimodif} 
leverages cross-modal differentiation to capture lip-speech 
inconsistencies; and UniCaCLF~\cite{yin2025context} employs context-aware contrastive learning. However, these SOTA methods 
predominantly rely on Transformer backbones. Despite their effectiveness, they suffer from quadratic computational complexity $O(T^2)$, creating significant efficiency bottlenecks for long video sequences.
\subsection{State Space Models}
State Space Models (SSMs) have evolved from early structured variants 
(e.g., S4~\cite{gu2021efficiently}, H3~\cite{fu2022hungry}, Gated State 
Space~\cite{mehta2022long}) to the recent Mamba~\cite{gu2023mamba} and 
Mamba2~\cite{dao2024transformers}, which achieve Transformer-level 
performance with linear complexity $O(T)$ via selective scanning. In 
the visual domain, architectures such as ViM~\cite{zhu2024vision}, 
VMamba~\cite{liu2024vmamba}, and Video Mamba Suite~\cite{chen2024video} have successfully adapted bidirectional scanning for image and video analysis, while MambaTAD~\cite{lu2025mambatad} explored temporal action localization. However, the application of Mamba to TFL remains largely unexplored. 
Crucially, existing visual Mamba architectures~\cite{zhu2024vision,
liu2024vmamba,chen2024video,lu2025mambatad} typically employ fixed 
dense scanning (stride=1), which limits their flexibility in handling multi-scale artifacts inherent in forgery localization. This gap motivates the development of adaptive scanning strategies tailored for TFL.

Recently, Mamba has been explored for multimodal tasks. 
AVS-Mamba~\cite{gong2025avs} and MUG~\cite{wang2025mug} introduce Mamba for audio-visual segmentation and parsing, respectively. 
Unlike our work which focuses on \textit{temporal forgery localization} via frequency-aware scanning, these methods primarily address spatial-temporal grouping for segmentation masks.
Similarly, LC-Mamba~\cite{jeong2025lc} explores local window shifts for frame interpolation. Our UniSkip-Mamba distinguishes itself by identifying the specific spectral redundancy in deepfakes and proposing \textit{skip-scanning} as a structural regularizer, rather than a generic efficiency booster.

\subsection{Multimodal Fusion in Deepfake Detection}
To detect sophisticated audio-visual inconsistencies (e.g., lip-speech 
desynchronization), leveraging multimodal cues is paramount. While 
early works explored late fusion, SOTA frameworks~\cite{
zhang2023ummaformer,yin2025context,bagchi2021hear,chugh2020not,
cai2022you} predominantly adopt early fusion via channel-wise 
concatenation. However, this approach inherently suffers from rigid 
frame alignment: it forces audio and visual features to align strictly 
at the same time step, restricting the model's ability to detect 
asynchronous artifacts where critical correlations between visual 
anomalies and auditory cues span across temporal offsets. Recent 
attempts to address this limitation remain constrained by the 
architectural assumptions of their underlying temporal backbones, 
highlighting the need for more flexible fusion strategies that can 
naturally accommodate asynchronous cross-modal correlations.

\begin{figure*}[t]
    \centering
    \includegraphics[width=\textwidth]{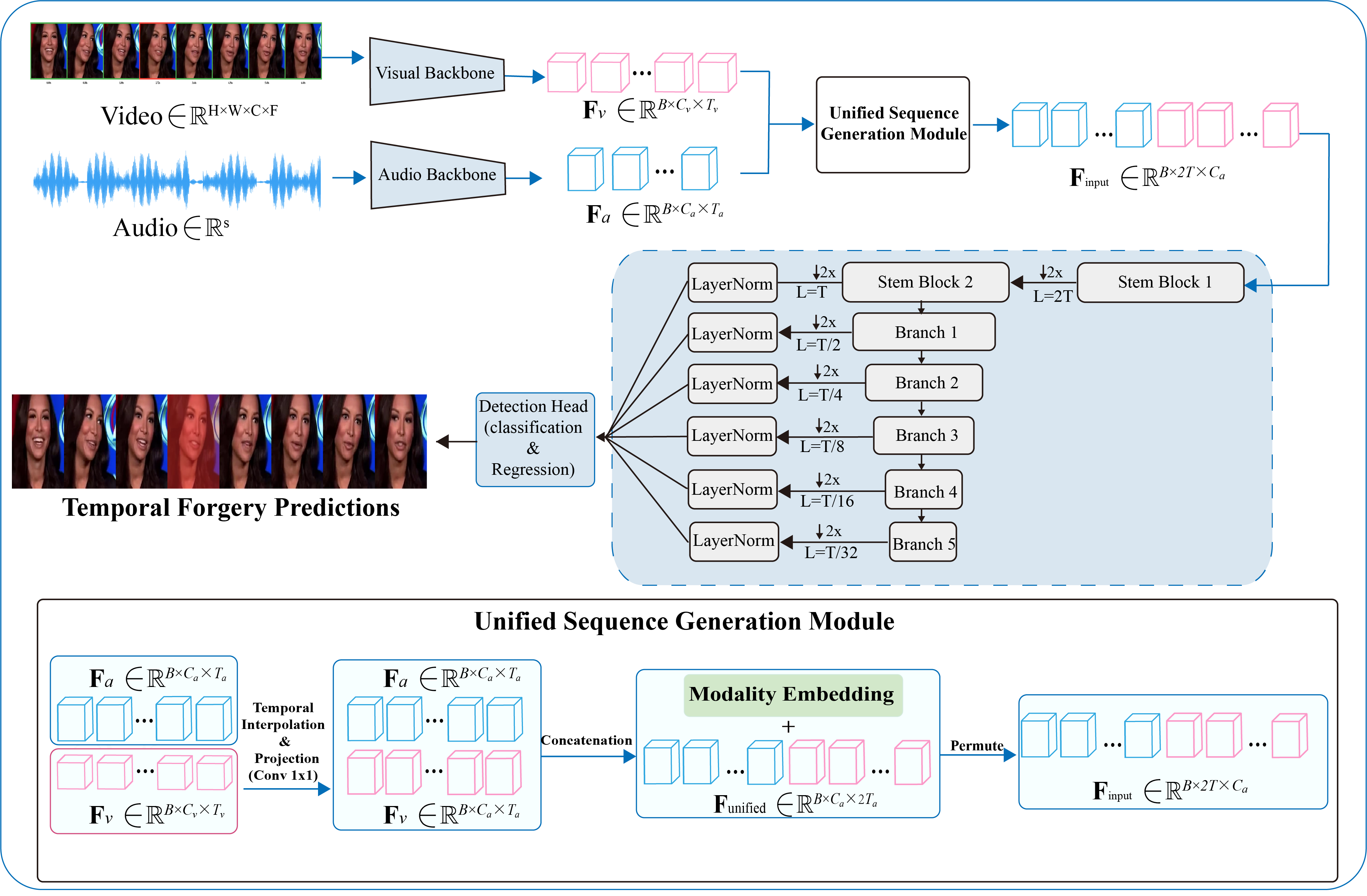}
    \caption{UniSkip-Mamba Framework Architecture. Given input video and audio, pre-trained backbones extract features that are fused via our Unified Sequence Generation Module—concatenating modalities temporally (not channel-wise) with learnable modality embeddings. The unified sequence $\mathbf{F}_{\text{input}} \in \mathbb{R}^{B \times 2T \times C_a}$ is processed through a hierarchical backbone with stem blocks and five progressive downsampling branches. Each branch stage concludes with a $2\times$ downsampling operation (indicated by $\downarrow 2\times$), progressively reducing the sequence length from $L=2T$ to $L=T/16$. Each stage employs Skip-Scanning Mamba blocks for efficient temporal modeling. The detection head produces final forgery localization predictions with precise temporal boundaries.}

    \label{fig:framework}
\end{figure*}

    
    
\section{Methodology}
The overall architecture of our proposed UniSkip-Mamba framework is illustrated in Fig.~\ref{fig:framework}. The framework consists of four key components: (1) feature extraction using pre-trained encoders, (2) unified multimodal sequence fusion with modality embeddings, (3) hierarchical skip-scanning Mamba backbone for efficient temporal modeling, and (4) detection head for boundary regression. We detail each component in the following subsections.

\subsection{Problem Formulation \& Preliminaries}
\label{sec:formulation}
\textbf{Problem Formulation.} Given an input video, we aim to localize forged segments $\mathcal{S} = \{(s_j, e_j, c_j)\}_{j=1}^{N}$, where $s_j, e_j, c_j$ denote the start time, end time, and forgery category, respectively. As shown in Fig.~\ref{fig:framework}, we first extract visual features $\mathbf{F}_v \in \mathbb{R}^{B \times C_v \times T}$ and audio features $\mathbf{F}_a \in \mathbb{R}^{B \times C_a \times T}$ using pre-trained encoders, where $B$ denotes batch size, $C_v$ and $C_a$ are feature dimensions, and $T$ is the temporal length. Our goal is to map these heterogeneous streams into a unified representation for precise boundary regression.

\textbf{Bidirectional Mamba Backbone.}
To efficiently model long-range dependencies, we leverage the \textbf{Mamba} architecture~\cite{gu2023mamba}, specifically the Decomposed Bidirectional Mamba (DBM) block~\cite{chen2024video}.
Unlike Transformers with $\mathcal{O}(T^2)$ complexity, Mamba employs a selective scan mechanism with linear complexity $\mathcal{O}(T)$. It discretizes the continuous state space equation $\mathbf{h}'(t) = \mathbf{A}\mathbf{h}(t) + \mathbf{B}x(t)$ into a recurrent form $\mathbf{h}_k = \bar{\mathbf{A}}\mathbf{h}_{k-1} + \bar{\mathbf{B}}x_k$, where system matrices are input-dependent.
To capture non-causal context essential for TFL, the DBM block processes sequences in both forward and backward directions, fusing the outputs to incorporate comprehensive temporal context. This serves as the fundamental building unit of our UniSkip-Mamba.

\subsection{Unified Multi-Modal Fusion}
\label{sec:umf}
To effectively model the cross-modal inconsistencies inherent in audio-visual forgeries, a principled fusion strategy is essential. As shown in the \textit{Unified Sequence Generation Module} of Fig.~\ref{fig:framework}, we transform the separate audio and visual streams into a single, modality-aware sequence.

First, assuming both modalities are synchronized to a sequence length $T$, we align the visual features $\mathbf{F}_v \in \mathbb{R}^{B \times C_v \times T}$ to the audio feature dimension $C_a$ using a linear projection layer:
\begin{equation}
    \mathbf{F}_v' = \text{Linear}_{C_v \rightarrow C_a}(\mathbf{F}_v) \in \mathbb{R}^{B \times C_a \times T}.
\end{equation}

Then, instead of channel-wise concatenation, we concatenate the aligned features along the temporal dimension to construct a unified sequence:
\begin{equation}
    \mathbf{F}_{\text{unified}} = \text{Concat}_{\text{temporal}}(\mathbf{F}_v', \mathbf{F}_a) \in \mathbb{R}^{B \times C_a \times 2T}.
\end{equation}

Finally, to explicitly inform the model of each token's origin, learnable modality embeddings are introduced. 
A modality ID sequence is created and passed through an embedding layer to generate the modality vectors $\mathbf{E}_{\text{modal}} \in \mathbb{R}^{B \times 2T \times C_a}$, which are then added element-wise to the unified sequence.
The final modality-aware sequence $\mathbf{F}_{\text{input}}$, which serves as the input to our backbone, is formulated as:
\begin{equation}
 \mathbf{F}_{\text{input}} = \text{Permute}(\mathbf{F}_{\text{unified}}) + \mathbf{E}_{\text{modal}} \in \mathbb{R}^{B \times 2T \times C_a},
\end{equation}
where $\text{Permute}(\cdot)$ transposes the last two dimensions to match the input requirements of the backbone.

This fusion method offers two key advantages. First, creating a single temporal sequence is naturally aligned with the architecture of our Mamba backbone, facilitating effective cross-modal temporal modeling. Second, the explicit modality embeddings provide a strong, differentiable prior, allowing the model to more efficiently learn forgery patterns instead of discovering modality information implicitly.
\begin{algorithm}[t]
\caption{Efficient Skip-Scanning Mamba (S-Mamba) Block}
\label{alg:s_mamba}
\begin{algorithmic}[1]
\Require Input sequence $\mathbf{X}_{in} \in \mathbb{R}^{B \times L \times C}$ (where $L=2T$ for the first layer), Skip step $p$, Bidirectional Mamba $\text{DBM}(\cdot)$
\Ensure Output sequence $\mathbf{X}_{out} \in \mathbb{R}^{B \times L \times C}$

\Function{Grouping}{$\mathbf{X}, p$}
    \State $B, L, C \gets \mathbf{X}.\text{shape}$
    \State $\mathbf{X} \gets \text{Reshape}(\mathbf{X}, (B, L/p, p, C))$ \Comment{Split sequence into $p$ interleaved groups}
    \State $\mathbf{X} \gets \text{Permute}(\mathbf{X}, (0, 2, 1, 3))$   \Comment{To $(B, p, L/p, C)$}
    \State $\mathbf{X} \gets \text{Reshape}(\mathbf{X}, (B \cdot p, L/p, C))$ \Comment{Stack groups into batch dim}
    \State \Return $\mathbf{X}$
\EndFunction

\Statex

\Function{Merging}{$\mathbf{X}, B, L, C, p$}
    \State $\mathbf{X} \gets \text{Reshape}(\mathbf{X}, (B, p, L/p, C))$
    \State $\mathbf{X} \gets \text{Permute}(\mathbf{X}, (0, 2, 1, 3))$   \Comment{To $(B, L/p, p, C)$}
    \State $\mathbf{X} \gets \text{Reshape}(\mathbf{X}, (B, L, C))$       \Comment{Restore original unified sequence}
    \State \Return $\mathbf{X}$
\EndFunction

\Statex
\State $\mathbf{X}_{norm} \gets \text{LayerNorm}(\mathbf{X}_{in})$
\State $\mathbf{X}_{grouped} \gets \text{\Call{Grouping}{$\mathbf{X}_{norm}, p$}}$ \Comment{Reduce effective length for efficiency}
\State $\mathbf{X}_{scanned} \gets \text{DBM}(\mathbf{X}_{grouped})$    \Comment{Process sub-sequences in parallel}
\State $\mathbf{X}_{merged} \gets \text{\Call{Merging}{$\mathbf{X}_{scanned}, B, L, C, p$}}$
\State $\mathbf{X}_{out} \gets \mathbf{X}_{in} + \text{Dropout}(\mathbf{X}_{merged})$ \Comment{Residual connection}
\State \Return $\mathbf{X}_{out}$
\end{algorithmic}
\end{algorithm}

\subsection{Efficient Skip-Scanning Mamba Block}
\label{ssec:smamba}

\begin{figure}[t]
    \centering
    \includegraphics[height=10cm, keepaspectratio]{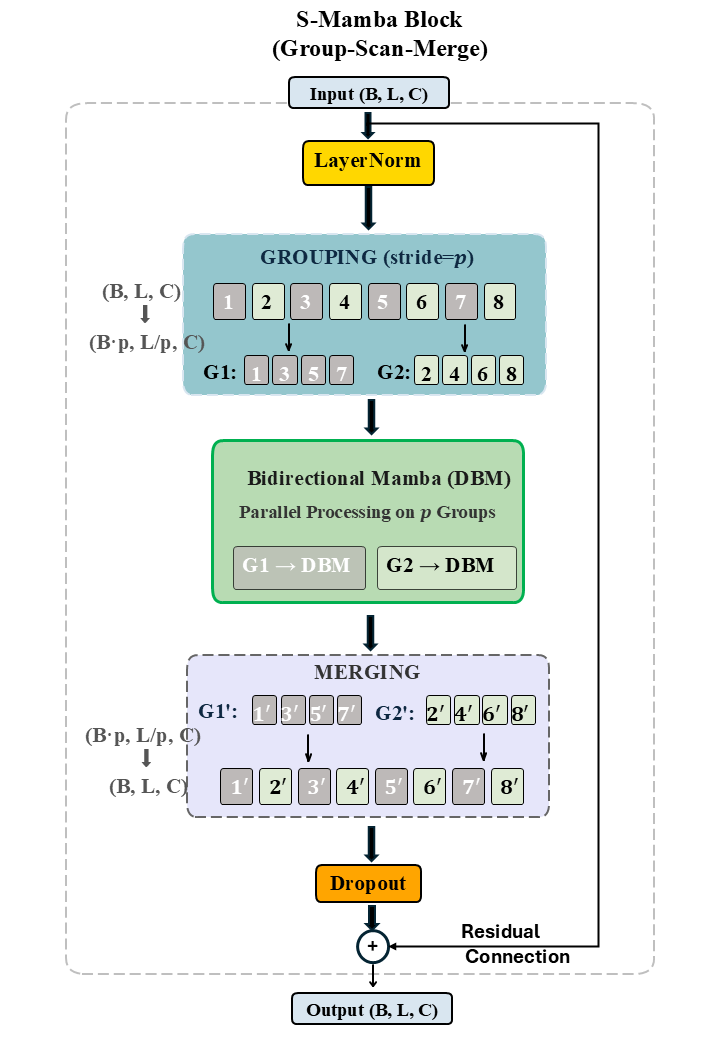}
    \caption{\textbf{Illustration of our proposed Efficient Skip-Scanning Mamba (S-Mamba) Block.} It introduces a ``Group-Scan-Merge'' mechanism: splitting the input sequence into $p$ groups to enable efficient parallel scanning with reduced sequence lengths, effectively acting as a temporal regularizer.}
    \label{fig:s_mamba}
\end{figure}
\textbf{Motivation: The Granularity--Efficiency Trade-off.}
Temporal forgery artifacts exhibit significant diversity in duration and scale. Micro-level artifacts require dense temporal sampling to capture subtle frame-level inconsistencies, whereas macro-level inconsistencies, such as prolonged audio-visual desynchronization, span much longer durations and primarily manifest in low and mid-frequency temporal patterns. A standard Mamba block scans the sequence step-by-step (stride 1). While this captures all temporal details, it is computationally redundant for detecting macro-level anomalies. More critically, dense scanning can lead to overfitting on high-frequency frame-level variations—including both genuine motion dynamics and compression artifacts—making the model vulnerable to noise in degraded data. To resolve this, we propose the \textbf{Skip-Scanning} mechanism, which introduces a configurable stride parameter $p$ to trade off temporal granularity for efficiency and long-range robustness. Our comprehensive frequency domain analysis (Section~\ref{ssec:freq_analysis}) validates this design, revealing that discriminative forgery patterns primarily reside in low and mid-frequency bands (normalized frequency 0-0.15), while high-frequency components often introduce noise that hinders robustness.

\textbf{Mechanism: Group-Scan-Merge.}
As detailed in Algorithm~\ref{alg:s_mamba}, our S-Mamba block processes the input sequence $\mathbf{X} \in \mathbb{R}^{B \times L \times C}$ through a three-stage pipeline:

    \paragraph{Grouping} Instead of scanning the sequence linearly, we first reshape and permute the input $\mathbf{X}$ into $p$ parallel sub-sequences. Mathematically, $\mathbf{X}$ is transformed into a tensor of shape $(B \cdot p) \times (L/p) \times C$. This operation effectively de-interleaves the original sequence, grouping tokens that are spaced $p$ steps apart.
    
    \paragraph{Parallel Scanning} These sub-sequences are then treated as independent batch elements and fed into a shared Bidirectional State Space Model. Specifically, we adopt the \textbf{Decomposed Bidirectional Mamba (DBM)} architecture~\cite{chen2024video}, which processes each sub-sequence in both forward and backward directions to ensure comprehensive temporal context awareness. By processing shorter sequences of length $L/p$, the SSM can propagate information over longer effective temporal distances with reduced computational overhead.
    
    \paragraph{Merging} Finally, the processed sub-sequences are interleaved back to the original order to restore the full temporal resolution $\mathbb{R}^{B \times L \times C}$. This ensures that the local temporal structure is preserved while global context has been integrated via the strided scan.

\subsection{Hierarchical Skip-Scanning Mamba Backbone}
\label{sec:backbone}

The unified, modality-aware sequence $\mathbf{F}_{input}$ is processed by our backbone. Unlike traditional dense-scanning approaches~\cite{zhu2024vision,liu2024vmamba,chen2024video,lu2025mambatad} that process every frame sequentially, we introduce a hierarchical architecture equipped with our novel \textbf{Efficient Skip-Scanning Mamba (S-Mamba)} blocks. This design is tailored to address the dual challenges of computational efficiency and multi-scale context modeling inherent in TFL.

\subsubsection{Hierarchical Architecture}
\label{ssec:hierarchical}

Our backbone follows a hierarchical design composed of two main stages: the \textit{Stem} and the \textit{Branch}.

\textbf{Stem.} The initial part of the backbone is the Stem, which operates on the full-resolution input sequence without downsampling. It consists of a series of S-Mamba blocks configured to extract fine-grained temporal features at the original sampling rate. This stage captures both low-frequency semantic patterns and mid-frequency dynamic transitions, which our frequency domain 
analysis~\ref{ssec:freq_analysis} reveals to be the most discriminative components for forgery detection. The dense sampling in the Stem is crucial for identifying subtle forgery artifacts, such as micro-expression inconsistencies or brief audio glitches, which require precise temporal alignment.

\textbf{Branch.} Following the Stem, the Branch consists of multiple stages. Each stage contains one or more S-Mamba blocks, but critically, each stage concludes with a downsampling operation that halves the temporal sequence length. This progressive downsampling structure allows the deeper layers of the backbone to possess a larger effective receptive field, enabling them to model longer-range temporal dependencies and capture coarse-grained, multi-scale contextual information essential for detecting sustained forgery patterns. The number of blocks in the Stem and in each Branch stage are configurable hyperparameters, allowing for flexible scaling of the model's capacity.
\subsection{Theoretical Analysis}
\label{subsec:theoretical_analysis}
We provide a theoretical foundation for our design choices from 
computational complexity, state-space dynamics, and signal processing perspectives. This analysis establishes \textit{a priori} design principles and testable predictions that are empirically validated in Section\ref{experiment}.

\subsubsection{Computational Complexity}

The standard Mamba scan has complexity $\mathcal{O}(L \cdot d_{\text{state}})$. Under skip-scanning with stride $p$, we partition the sequence into $p$ groups of length $L/p$, processed in parallel. While the asymptotic complexity remains $\mathcal{O}(L \cdot d_{\text{state}})$, the reduction in sequential dependency 
depth from $L$ to $L/p$ trades sequential overhead for parallel width.

\textbf{Prediction (Efficiency):} On modern GPUs limited by sequential recurrence latency, stride-$p$ scanning should yield inference speedups approaching linear scaling with $p$ (e.g., $\sim 2\times$ for $p=2$).

\subsubsection{Unified Sequence as Phase Preserver}

From a signal processing perspective, cross-modal forgery detection 
requires modeling \textit{phase relationships} (e.g., audio-visual 
synchronization) between modalities. Channel-wise fusion $\mathbf{x}_t = [\mathbf{v}_t; \mathbf{a}_t]$ forces rigid temporal alignment, structurally constraining correlation modeling to local windows $|\Delta t| \leq R$ (where $R$ is the effective receptive field). Thus, long-range desynchronization ($\Delta t > R$) 
becomes theoretically undetectable.

In contrast, our unified sequence $[\mathbf{V}_1, \ldots, \mathbf{V}_T, \mathbf{A}_1, \ldots, \mathbf{A}_T]$ converts cross-modal correlation into a \textit{long-range dependency problem}. The SSM can model correlations between $\mathbf{v}_t$ and $\mathbf{a}_{t+k}$ for arbitrary temporal lags $k$ within a single forward pass.

\textbf{Prediction (Phase Modeling):} The unified representation should exhibit superior capability in detecting asynchronous forgeries with large temporal offsets compared to channel-wise baselines, which are structurally bound by local alignment constraints.

\subsubsection{Skip-Scanning: Frequency-Aware Regularization}

\textbf{Mechanism Clarification.}
Skip-scanning via Group-Scan-Merge is architecturally distinct from 
pooling or downsampling: it \textit{retains representations for all tokens} while modifying the scanning topology. We analyze this as structured regularization affecting the model's learned inductive bias.

\textbf{State-Space Perspective.}
Under stride-$p$ scanning, the recurrence within each group operates with effective state transition matrix $\bar{\mathbf{A}}_{\text{skip}} = \bar{\mathbf{A}}^p$. 
Since $\bar{\mathbf{A}} = \exp(\Delta \mathbf{A})$ is typically contractive (i.e., $\|\bar{\mathbf{A}}\| < 1$ for well-trained SSMs), we have $\|\bar{\mathbf{A}}^p\| = \|\bar{\mathbf{A}}\|^p$, which decays exponentially with $p$. For instance, if $\|\bar{\mathbf{A}}\| = 0.8$, then stride-2 yields $\|\bar{\mathbf{A}}^2\| = 0.64$ (20\% reduction) and stride-4 yields 
$0.41$ (49\% reduction). This exponential scaling implies that skip-scanning \textbf{accelerates the decay of high-frequency state dynamics}, biasing the model to down-weight short-term temporal jitter.

\textbf{Frequency-Domain Analysis \& Optimal Stride Selection.}
The effective temporal sampling rate within each group is reduced by factor $p$, yielding an \textit{approximate} effective Nyquist frequency $f_N^{\text{eff}}(p) \approx 1/(2p)$ (normalized). Crucially, this is a functional approximation rather than physical downsampling, as our Group-Scan-Merge mechanism retains all tokens via parallel processing. The Nyquist-like constraint emerges specifically from the reduced recurrence depth within each group, which effectively limits the upper bound of temporal frequencies that can be captured during the scanning process.

Based on this spectral property, we consider the hypothesis that discriminative forgery patterns—such as semantic-level inconsistencies, prosodic mismatches, and audio-visual synchronization drift—are inherently low/mid-frequency temporal phenomena. This hypothesis is biologically motivated by human perception, where the detection of semantic discontinuities and synchronization errors typically operates at timescales of 50--200ms. In the context of feature sequences extracted at 6.25Hz (our AV-Deepfake1M setting), these physical timescales map to normalized frequencies of approximately 0.05--0.2. Consequently, we posit that the critical discriminative band satisfies $f_d \lesssim 0.2$.

Under this hypothesis, we can theoretically rank stride choices:
\begin{itemize}
    \item \textbf{Stride 2} ($f_N^{\text{eff}} \approx 0.25$): Optimal. The discriminative bandwidth ($f_d \lesssim 0.2$) is fully preserved with a safety margin, while higher-frequency noise is naturally attenuated.
    \item \textbf{Stride 4} ($f_N^{\text{eff}} \approx 0.125$): Aggressive. It risks attenuating discriminative cues if $f_d$ approaches or exceeds $0.125$, potentially causing information loss.
\end{itemize}

\textbf{Prediction (Frequency Selectivity):} If our hypothesis holds, then:
\begin{enumerate}
    \item Removing low/mid-frequency components ($f < f_d$) should cause severe performance degradation across all models.
    \item Removing high-frequency components ($f >f_N^{\text{eff}}$) should have a minimal or even positive impact on stride-$p$ models, as these frequencies are already down-weighted by the architecture.
    \item Stride-4 should exhibit increased sensitivity to mid-frequency removal compared to stride-2, as $f_N^{\text{eff}}(4) < f_d$.
\end{enumerate}

\textbf{Prediction (Robustness):} Models with larger strides ($p > 1$) 
possess a low-pass inductive bias. They should exhibit superior robustness to data perturbations that primarily corrupt high-frequency details (e.g., compression artifacts, additive noise, blur) compared to the dense-scanning baseline.

\textbf{Soft Regularization Paradigm.} Critically, our mechanism differs from conventional hard-filtering. Rather than discarding high-frequency components at the input level, skip-scanning implements a form of \textit{soft structural regularization}. By altering the scanning topology, we introduce an inductive bias that accelerates the decay of state dynamics, as mathematically derived in the Supplementary Material. This effectively biases the \textit{learning process} toward robust, low-frequency temporal patterns. Importantly, the Group-Scan-Merge architecture retains the full \textit{representational capacity} to encode fine-grained details via all-token retention. This design enables the model to prioritize stable semantic features for detection while adaptively attending to high-frequency cues only when explicitly required by the loss function, such as for precise boundary refinement.
\section{Experimental}
\label{experiment}
\subsection{Datasets}
We conduct extensive experiments on two challenging datasets, covering a wide spectrum of forgery scenarios.

\begin{itemize}
    \item \textbf{LAV-DF~\cite{cai2023glitch}}: A large-scale multimodal forgery dataset comprising 136,304 videos. It features four data types: pristine, visual-only tampering, audio-only tampering, and audio-visual tampering. The forged segments are typically short, with durations ranging from 0 to 1.6 seconds.
    
    \item \textbf{AV-Deepfake1M~\cite{cai2024av}}: The largest multimodal forgery dataset. To ensure the validity of audio-visual temporal localization, we excluded silent videos, resulting in a refined training set of 1,091,420 videos and a validation set of 56,907 videos. We use the validation set for testing as the official test set metadata is withheld.
\end{itemize}

\subsection{Implementation Details}
\label{ssec:implementation}
\textbf{Feature Extraction.} 
Following previous works~\cite{zhang2023ummaformer}, we utilize pre-extracted features to ensure a fair comparison.
For the \textbf{LAV-DF~\cite{cai2023glitch}} dataset, we use the two-stream TSN network pretrained on ActivityNet~\cite{caba2015activitynet} to extract visual features (4096-dim), combining RGB and optical flow. For audio, we employ a pre-trained BYOL-A model~\cite{niizumi2021byol} trained on AudioSet~\cite{gemmeke2017audio}, yielding 2048-dim features.
For the \textbf{AV-Deepfake1M~\cite{cai2024av}} dataset, we adopt more advanced feature extractors: VideoMAE V2-Small~\cite{wang2023videomae} (distilled from Giant) is used for visual features (384-dim), and Wav2Vec 2.0~\cite{baevski2020wav2vec} (XLS-R-300M) is used for audio features (1024-dim).

\subsubsection{Model Configuration.} 
Our hierarchical Mamba backbone is configured with an architecture of $[2, 2, 5]$ blocks for the embedding, stem, and branch stages, respectively. 
The hidden dimension $C$ is set to 512.
We investigate skip-scanning strides of $p \in \{1, 2, 4\}$ to analyze the trade-off between granularity and efficiency, where $p=1$ represents dense scanning (Full Scan).

\subsubsection{Training Settings.} 
The model is implemented in PyTorch 2.1.2 with CUDA 11.8. 
We train the network end-to-end using the AdamW optimizer with an initial learning rate of $1 \times 10^{-4}$ and weight decay of $0.05$. 
A cosine annealing schedule with warmup is employed. 
Training runs for 50 epochs with a batch size of 16. 
All experiments are conducted on NVIDIA Tesla V100-SXM2 (32GB) GPUs.

\begin{table*}[htbp]
\centering
\caption{Temporal forgery localization results on LAV-DF~\cite{cai2023glitch}. Modality denotes the model's input type with $\mathcal{V}$ being visual and $\mathcal{A}$ audio. \textbf{Bold} indicates best and \underline{underline} second to best performance. We report mAP as the average of AP at different thresholds.}
\label{tab:results}
\setlength{\tabcolsep}{6pt} 
\begin{tabular}{l c | c c c c | c c c c} 
\toprule
Method & Modality & AP@0.5 & AP@0.75 & AP@0.95 & \textbf{Avg.} & AR@100 & AR@50 & AR@20 & AR@10 \\
\midrule

MDS~\cite{chugh2020not}  & $\mathcal{AV}$ & 12.8 & 1.6 & 0.0 & 4.8 & 37.9 & 36.7 & 34.4 & 32.2 \\
AVFusion~\cite{bagchi2021hear}  & $\mathcal{AV}$ & 65.4 & 23.9 & 0.1 & 29.8 & 63.0 & 59.3 & 54.8 & 52.1 \\
BA-TFD~\cite{cai2022you}  & $\mathcal{AV}$ & 76.9 & 38.5 & 0.3 & 38.6 & 66.9 & 64.1 & 60.8 & 58.4 \\
BA-TFD+~\cite{cai2023glitch} & $\mathcal{AV}$ & 96.3 & 85.0 & 4.4 & 61.9 & 81.6 & 80.5 & 79.4 & 78.8 \\
TriDet~\cite{shi2023tridet}  & $\mathcal{V}$ & 96.3 & 86.8 & 23.6 & 68.9 & 91.0 & 90.4 & 89.7 & 88.7 \\
ActionMamba~\cite{wen2025actionmamba}  & $\mathcal{V}$ & 97.6 & 94.6 & 39.1 & 77.1 & 85.1 & 85.0 & 84.9 & 84.5 \\
UMMAFormer~\cite{zhang2023ummaformer} & $\mathcal{AV}$ & \textbf{98.8} & 95.5 & 37.6 & 77.3 & 92.4 & 92.5 & 92.5 & 92.1 \\
DiMoDif~\cite{koutlis2024dimodif} & $\mathcal{AV}$ & 95.5 & 87.9 & 20.6 & 68.0 & 94.2 & \underline{93.7} & 92.7 & 91.4 \\

UniCaCLF~\cite{yin2025context} & $\mathcal{AV}$ & 97.8 & 93.1 & 53.6 & 81.5 & \textbf{94.6} & \textbf{94.0} & \textbf{93.8} & \textbf{93.2} \\
\midrule
UniSkip-Mamba (stride=1) & $\mathcal{AV}$ & 98.5 & \underline{96.7} & \underline{63.0} & \underline{86.0} & \underline{94.3} & 93.6 & \underline{92.9} & \underline{92.3} \\
UniSkip-Mamba (stride=2) & $\mathcal{AV}$ & \underline{98.6} & \textbf{97.1} & \textbf{63.4} & \textbf{86.2} & 93.7 & 92.6 & 91.7 & 91.3 \\
\bottomrule
\end{tabular}
\end{table*}

\begin{table*}[htbp]
\centering
\caption{Temporal forgery localization results on AV-Deepfake1M~\cite{cai2024av}. Modality denotes the model's input type with $\mathcal{V}$ being visual and $\mathcal{A}$ audio. \textbf{Bold} indicates best and \underline{underline} second to best performance. * Reports validation performance.}
\label{tab:results_av_deepfake1m}
\setlength{\tabcolsep}{4pt} 
\begin{tabular}{l c | c c c c c | c c c c}
\toprule
Method & Modality & AP@0.5 & AP@0.75 & AP@0.9 & AP@0.95 & \textbf{Avg.} & AR@50 & AR@20 & AR@10 & AR@5 \\
\midrule
MesoInception4~\cite{afchar2018mesonet} & $\mathcal{V}$ & 08.50 & 05.16 & 01.89 & 00.50 & 04.01 & 39.27 & 39.00 & 35.78 & 24.59 \\
ActionFormer+VideoMAEv2~\cite{zhang2022actionformer,wang2023videomae} & $\mathcal{V}$ & 20.24 & 05.73 & 00.57 & 00.07 & 06.65 & 19.97 & 19.81 & 19.11 & 17.80 \\
BA-TFD~\cite{cai2022you} & $\mathcal{AV}$ & 37.37 & 06.34 & 00.19 & 00.02 & 10.98 & 45.55 & 35.95 & 30.66 & 26.82 \\
BA-TFD+~\cite{cai2023glitch} & $\mathcal{AV}$ & 44.42 & 13.64 & 00.48 & 00.03 & 14.64 & 48.86 & 40.37 & 34.67 & 29.88 \\
UMMAFormer~\cite{zhang2023ummaformer} & $\mathcal{AV}$ & 51.64 & 28.07 & 07.65 & 01.58 & 22.24 & 44.07 & 43.45 & 42.09 & 40.27 \\
DiMoDif~\cite{koutlis2024dimodif} & $\mathcal{AV}$ & 86.93 & 75.95 & 28.72 & 05.43 & 49.26 & \textbf{81.57} & \textbf{80.25} & \textbf{78.84} & \textbf{76.64} \\
\midrule
UniSkip-Mamba (stride=1) & $\mathcal{AV}$ & 87.2 & 75.3 & 43.7 & 22.0 & 57.1 & 78.1 & 75.9 & 73.7 & 69.7 \\

UniSkip-Mamba (stride=2) & $\mathcal{AV}$ & \textbf{90.60} & \textbf{81.80} & \textbf{52.70} & \textbf{29.20} & \textbf{63.58} & \underline{80.90} & \underline{79.30} & \underline{77.50} & \underline{74.70} \\
UniSkip-Mamba (stride=4) & $\mathcal{AV}$ & \underline{90.50} & \underline{80.70} & \underline{50.60} & \underline{27.60} & \underline{62.35} & 78.80 & 77.20 & 75.50 & 72.80 \\
\bottomrule
\end{tabular}
\end{table*}

\subsection{Main Results and Analysis}
\label{ssec:main_results}

Tables~\ref{tab:results} and~\ref{tab:results_av_deepfake1m} present comprehensive comparisons with SOTA methods on LAV-DF~\cite{cai2023glitch} and AV-Deepfake1M~\cite{cai2024av}. Our analysis reveals several key findings demonstrating the effectiveness of UniSkip-Mamba.

\subsubsection{Performance on LAV-DF}

On LAV-DF~\cite{cai2023glitch}, our stride 2 model achieves \textbf{63.4\% AP@0.95}, representing a \textbf{+9.8\% absolute improvement} over the previous best method UniCaCLF~\cite{yin2025context} (53.6\%). This substantial gain at the most stringent IoU threshold demonstrates that skip-scanning enables exceptionally precise boundary localization. The improvement is even more pronounced compared to UMMAFormer~\cite{zhang2023ummaformer} (+25.8\%) and DiMoDif~\cite{koutlis2024dimodif} (+42.8\%). Our method also achieves \textbf{97.1\% AP@0.75} (+4.0\% over UniCaCLF) and \textbf{86.2\% average mAP} (+4.7\% over UniCaCLF), demonstrating consistent superiority across all IoU thresholds.

Comparing scanning strides, stride 2 (63.4\% AP@0.95) slightly outperforms stride 1 (63.0\%), confirming that temporal subsampling acts as beneficial regularization. Combined with stride 2's 2$\times$ inference speedup (Section~\ref{ssec:efficiency}), this establishes it as the optimal configuration.

\subsubsection{Performance on AV-Deepfake1M}

On the larger AV-Deepfake1M dataset~\cite{cai2024av}, UniSkip-Mamba demonstrates even more substantial improvements. Our stride 2 model achieves remarkable gains over DiMoDif~\cite{koutlis2024dimodif}: \textbf{+3.67\% AP@0.5}, \textbf{+5.85\% AP@0.75}, \textbf{+23.98\% AP@0.9}, \textbf{+23.77\% AP@0.95}, and \textbf{+14.32\% average mAP}. The progressively larger gains at higher IoU thresholds reveal a critical advantage: while DiMoDif detects forgeries at a coarse level, our method excels at pinpointing exact temporal boundaries.

Compared to UMMAFormer~\cite{zhang2023ummaformer}, the improvement is even more striking: \textbf{+27.62\% AP@0.95} and \textbf{+41.34\% average mAP}. This validates our hypothesis that Mamba's linear complexity enables effective long-range temporal modeling, whereas Transformer's quadratic complexity forces compromises in sequence length or model capacity.

Notably, stride 2 shows a substantial advantage over stride 1 on this larger dataset (+6.48\% mAP vs. +0.2\% on LAV-DF), suggesting that skip-scanning's regularization effect becomes more pronounced on diverse, large-scale data where models are prone to overfitting high-frequency noise. Stride 4 (62.35\% mAP) also significantly outperforms stride 1 (57.1\% mAP), confirming that temporal subsampling consistently improves generalization.

\subsubsection{Key Insights}

\textbf{Skip-Scanning as Regularization.} The consistent performance gains of stride 2 over stride 1, particularly on large-scale data, provide strong empirical evidence that skip-scanning acts as structural regularization. Our frequency domain analysis (Section~\ref{ssec:freq_analysis}) provides theoretical justification, showing that stride 2 naturally attenuates high-frequency noise while preserving discriminative low and mid-frequency patterns.

\textbf{Mamba vs. Transformer.} The dramatic improvements over UMMAFormer~\cite{zhang2023ummaformer} demonstrate that Mamba's linear complexity fundamentally enables better long-range temporal modeling for forgery detection, eliminating the sequence length vs. model capacity trade-off inherent in Transformers.

\textbf{Multimodal Advantage.} Our method significantly outperforms DiMoDif~\cite{koutlis2024dimodif} (+23.77\% AP@0.95), which also uses audio-visual information through specialized speech models. Our unified sequence representation with modality embeddings provides a more flexible framework that captures diverse cross-modal discrepancies beyond lip-sync errors, including temporal desynchronization, prosodic inconsistencies, and quality mismatches.

In summary, UniSkip-Mamba establishes new SOTA results on both benchmarks, with particularly dramatic improvements at high-precision metrics. The consistent superiority of stride 2 validates our core innovation: skip-scanning is not a compromise for efficiency, but a principled regularization strategy that improves both accuracy and speed.

\subsection{Robustness Evaluation}
\label{ssec:robustness}

To comprehensively evaluate real-world stability, we conduct robustness tests under various cross-modal perturbations on 1,000 sampled videos from AV-Deepfake1M~\cite{cai2024av} validation set. We report \textbf{mAP@0.95}, the most rigorous metric for precise boundary localization under degradation, comparing UniSkip-Mamba stride $\in \{1, 2, 4\}$ against SOTA method DiMoDif~\cite{koutlis2024dimodif}.

\begin{figure*}[t]
    \centering
    \includegraphics[width=\textwidth]{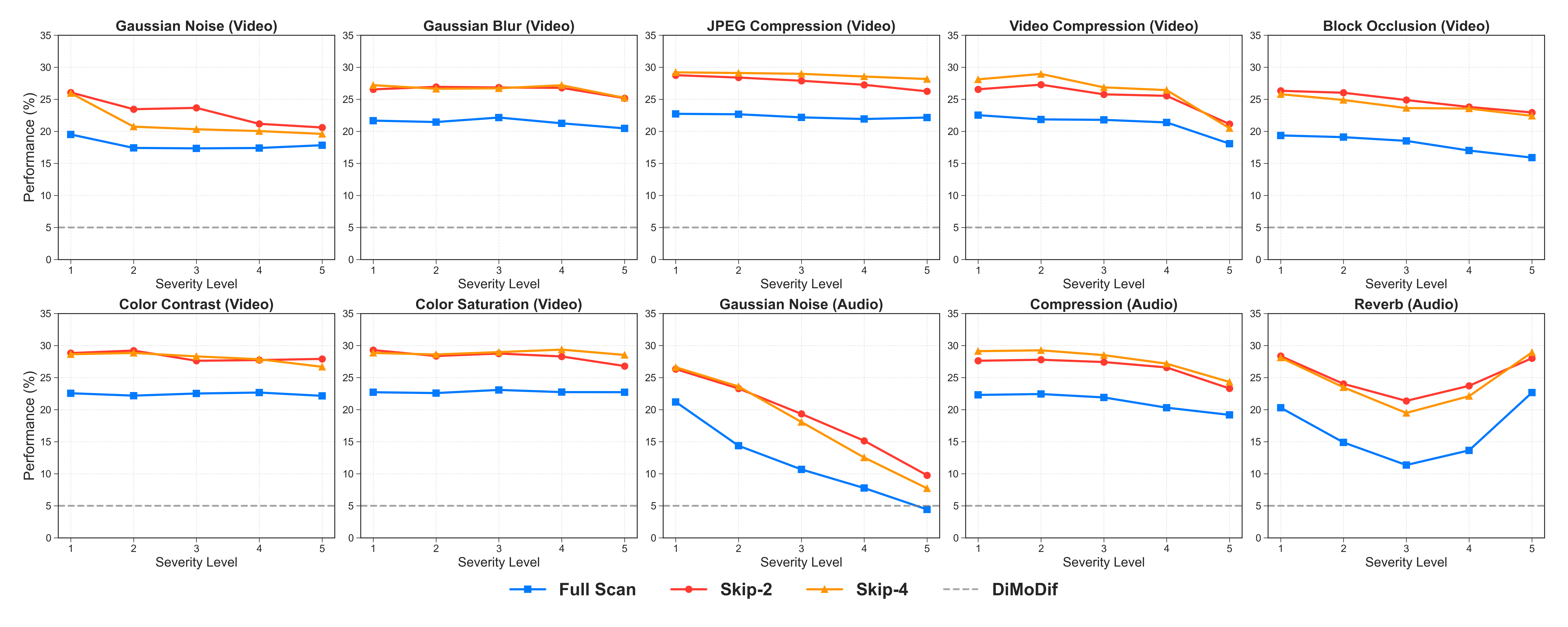}
    \caption{\textbf{Robustness evaluation under cross-modal perturbations.} We report mAP@0.95 across 5 intensity levels for 10 degradation types (7 visual, 3 audio). \textbf{(a)-(g)} Visual perturbations: Gaussian noise, blur, compression, occlusion, saturation, contrast. \textbf{(h)-(j)} Audio perturbations: noise, compression, reverberation. Skip-scanning models (stride $>1$) consistently outperform dense scanning (stride 1), demonstrating superior resilience. DiMoDif~\cite{koutlis2024dimodif} achieves $<5\%$ mAP across all settings (omitted for clarity).}
    \label{fig:robustness}
\end{figure*}

\textbf{Perturbation Settings.} We applied 10 types of degradations with 5 intensity levels. Detailed parameter settings are provided in the Supplementary Material.
\textbf{Results Analysis.} Figure~\ref{fig:robustness} reveals three critical findings:

\textbf{(1) Skip-Scanning Provides Consistent Robustness Gains.} Across all 10 perturbation types, stride 2 and stride 4 consistently outperform stride 1 by \textbf{5-8\% absolute mAP}. This validates our hypothesis that temporal subsampling acts as frequency-aware regularization—reducing sensitivity to high-frequency perturbations while preserving discriminative low/mid-frequency patterns (Section~\ref{ssec:freq_analysis}). The natural filtering effect prevents performance degradation when high-frequency details are corrupted.

\textbf{(2) Superior Resilience Under Severe Degradation.} At extreme perturbation levels (Block Occlusion level 5, Video Compression CRF 51), stride 1 degrades catastrophically to 15.9\% and 18.1\%, while stride 4 maintains 22.4\% and 20.5\%—demonstrating \textbf{1.4$\times$ better resilience}. The sparse scanning mechanism effectively bypasses local corruptions and reconstructs forgery context from remaining reliable frames. Similarly, under intense audio noise (level 5), stride 1 drops to 4.43\% while stride 2 retains 9.76\%, \textbf{doubling robustness} by filtering transient noise and focusing on underlying prosodic inconsistencies.

\textbf{(3) Dramatic Advantage Over SOTA.} DiMoDif~\cite{koutlis2024dimodif} achieves $<5\%$ mAP@0.95 across all conditions, highlighting the extreme difficulty of precise boundary localization under degradation. Our framework's unique capability stems from frequency-principled design: stride=2's Nyquist frequency (0.25 Hz) acts as a natural barrier—perturbations corrupting high-frequency components ($>0.25$ Hz) cannot affect the model's decision-making process, providing \textit{provable} robustness grounded in signal processing theory.

\begin{figure*}[t]
    \centering
    
    \subfloat[Training Efficiency]{%
        \includegraphics[width=0.32\textwidth]{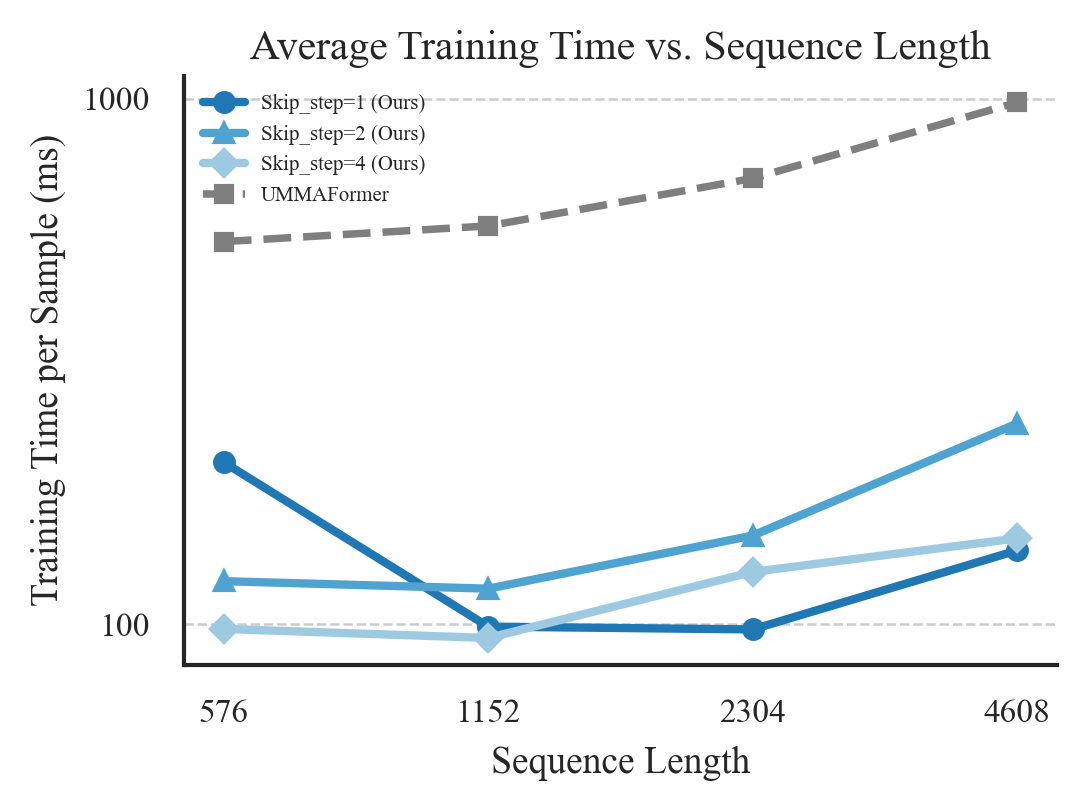}%
        \label{fig:eff_train}}
    \hfill
    \subfloat[Inference Latency]{%
        \includegraphics[width=0.32\textwidth]{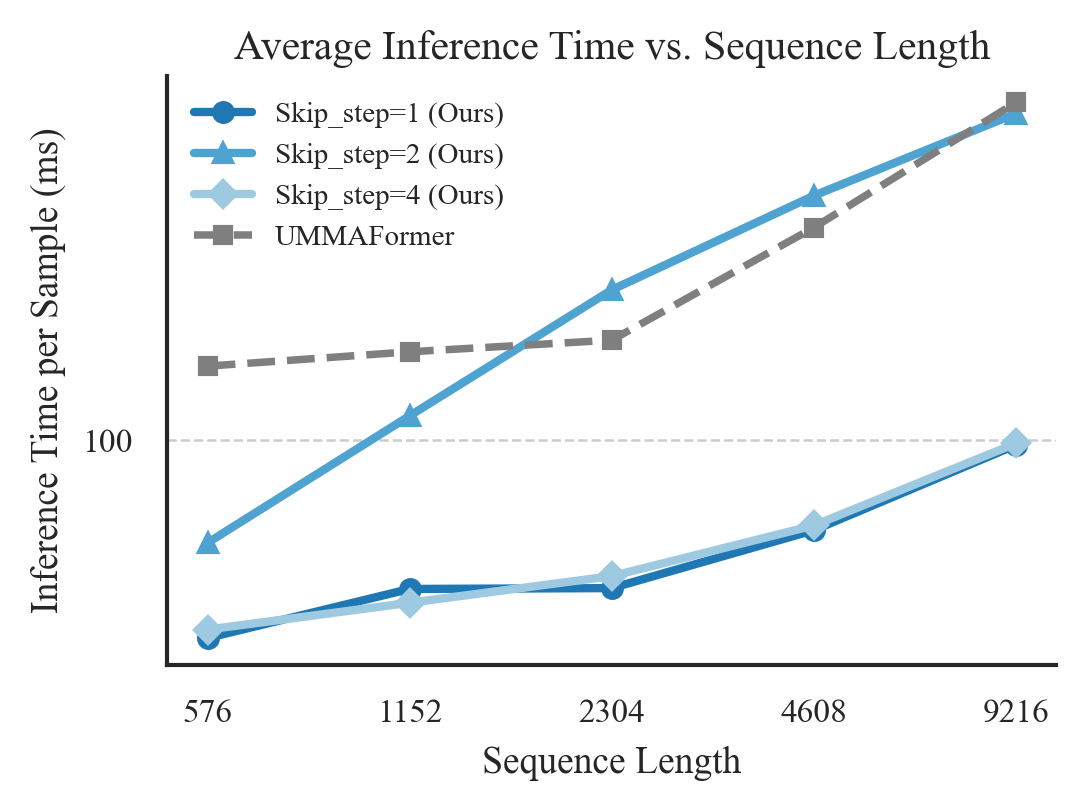}%
        \label{fig:eff_infer}}
    \hfill
    \subfloat[Batch Scalability]{%
        \includegraphics[width=0.32\textwidth]{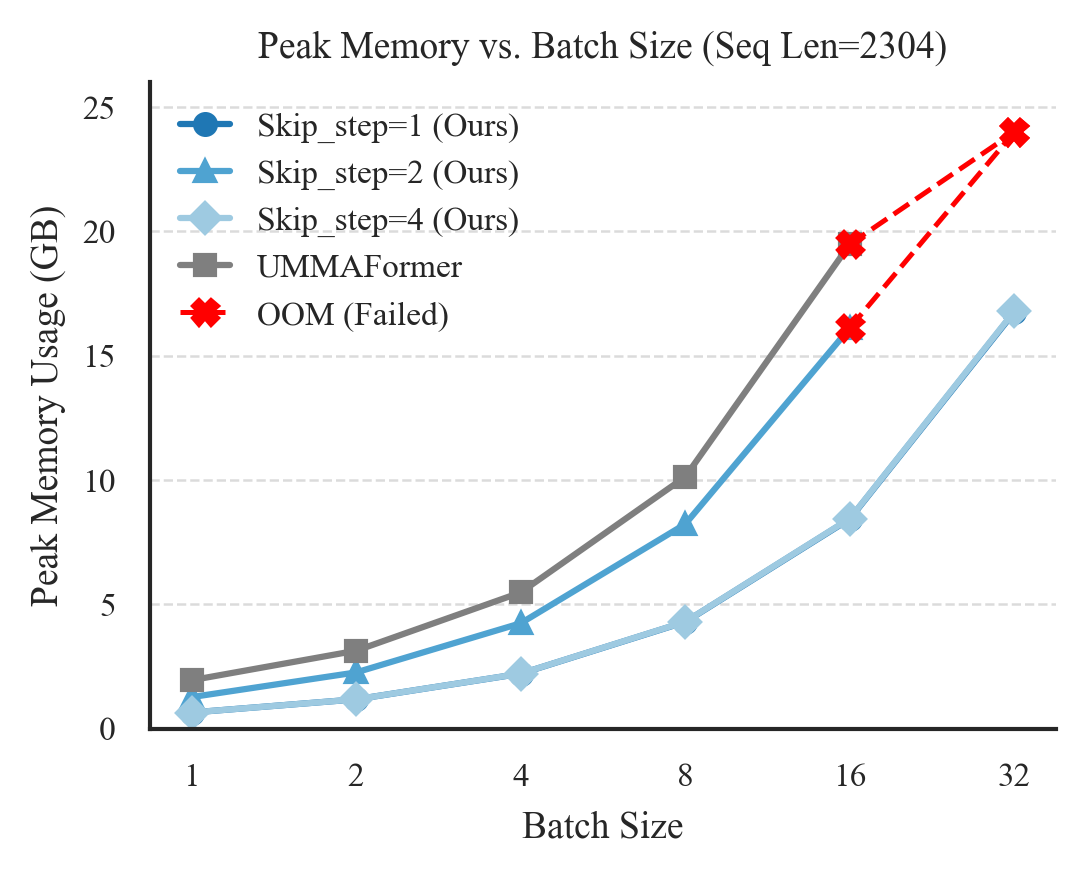}%
        \label{fig:eff_scale}}
    
    \caption{\textbf{Comprehensive Efficiency Analysis.} (a) Comparison of training time and memory usage across sequence lengths. (b) Inference latency on ultra-long sequences. (c) Scalability test showing peak memory and OOM thresholds under increasing batch sizes. Our UniSkip-Mamba demonstrates superior speed and scalability compared to the UMMAFormer ~\cite{zhang2023ummaformer} baseline.}
    \label{fig:efficiency_analysis}
\end{figure*}

\subsection{Computational Efficiency and Scalability Analysis}
\label{ssec:efficiency}

To demonstrate the practical superiority of our proposed \textbf{Uni-Mamba} framework over traditional Transformer-based methods, we conducted a comprehensive evaluation of computational efficiency. We selected UMMAFormer~\cite{zhang2023ummaformer}, a SOTA Transformer-based TFL model, as the baseline. The comparison focuses on three critical dimensions: Training Efficiency, Inference Latency, and Batch Scalability. All experiments were performed on a single NVIDIA Tesla V100-SXM2-32GB GPU.

\textbf{Training Efficiency: Speed vs. Memory.}
We first evaluated the training time and GPU memory consumption across varying input sequence lengths ($T \in [576, 4608]$). As illustrated in Fig.~\ref{fig:eff_train}, our method consistently outperforms UMMAFormer~\cite{zhang2023ummaformer} in terms of speed. Notably, the \textbf{stride 1} variant achieves a speedup of \textbf{3$\times$ to 5$\times$} compared to the baseline. For instance, at $T=4608$, our model completes a forward-backward pass in just \textbf{138ms}, whereas UMMAFormer requires \textbf{983ms}. This significant reduction in training latency allows for much faster experimental iteration. While our method exhibits a slightly higher memory footprint at extreme sequence lengths (e.g., 1.67GB vs. 0.85GB at $T=4608$) due to the bidirectional state expansion, this cost is well within the capacity of modern GPUs and is justified by the substantial gain in training speed.

\textbf{Inference Latency and Real-time Capability.}
In real-world forensic applications, low latency is paramount. We measured the inference time across sequence lengths up to 9216 frames. As shown in Fig.~\ref{fig:eff_infer}, UMMAFormer ~\cite{zhang2023ummaformer} suffers from a quadratic increase in latency, reaching \textbf{600ms} at $T=9216$, which hinders real-time processing. In stark contrast, our \textbf{stride 1} model maintains an impressively low latency of \textbf{97.4ms} at the same length, representing a \textbf{6$\times$ speedup}. Even the \textbf{stride 2} variant remains significantly faster than the baseline, ensuring stability for long-form video analysis.

\textbf{Scalability: Batch Size and Throughput.}
Finally, we assessed the model's scalability by increasing the batch size until Out-Of-Memory (OOM) occurred, with sequence length fixed at $T=2304$. The results, presented in Fig.~\ref{fig:eff_scale}, highlight the superior throughput of our approach. UMMAFormer ~\cite{zhang2023ummaformer} encounters OOM at a batch size of \textbf{32}, limiting its maximum viable batch size to \textbf{16}. Conversely, our \textbf{stride 1} and \textbf{stride 4} variants scale efficiently up to a batch size of \textbf{56}, crashing only at 64. This represents a \textbf{3.5$\times$ improvement} in parallel processing capacity. Furthermore, at a batch size of 1, our method occupies only \textbf{0.653 GB} of video memory, compared to \textbf{1.951 GB} for UMMAFormer. Consequently, our method achieves a maximum throughput of approximately \textbf{25 samples/sec}, far exceeding the \textbf{7.3 samples/sec} of the baseline.

In conclusion, these results demonstrate that UniSkip-Mamba addresses the critical efficiency bottlenecks of existing TFL frameworks. By offering high throughput, low latency, and excellent scalability, it provides a practical and robust solution for large-scale multimedia forensics.
\begin{table}[t]
    \centering
    \caption{Comprehensive ablation study on LAV-DF. We systematically evaluate 
    fusion strategies, modality embeddings, skip-scanning, and critically, the 
    backbone-fusion compatibility. Results demonstrate that unified sequence 
    requires architectural synergy with Mamba's linear complexity and 
    position-agnostic state propagation.}
    \label{tab:ablation_full}
    
    \resizebox{\columnwidth}{!}{%
        \setlength{\tabcolsep}{4pt}
        \begin{tabular}{l c c c c}
            \toprule
            \textbf{Method} & \textbf{Fusion} & \textbf{Mod. Emb.} & \textbf{Skip Scan} & \textbf{AP@0.95} \\
            
            \midrule
            \multicolumn{5}{l}{\textit{Transformer Backbone}} \\
            UMMAFormer~\cite{zhang2023ummaformer} & Channel & $\times$ & $\times$ & 37.6 \\
            UMMAFormer + Unified & Unified & \checkmark & $\times$ & 34.6 \\
            
            \midrule
            \multicolumn{5}{l}{\textit{Mamba Backbone (Ours)}} \\
            ActionMamba~\cite{wen2025actionmamba} & Channel & $\times$ & $\times$ & 39.1 \\
            Variant A & Channel & $\times$ & \checkmark & 36.7 \\ 
            Variant B & Channel & \checkmark & $\times$ & 34.7 \\ 
            \cmidrule(lr){1-5}
            Variant C & Unified & $\times$ & $\times$ & 43.3 \\
            UniSkip-Mamba (full-scan) & Unified & \checkmark & $\times$ & 63.0 \\ 
            \textbf{UniSkip-Mamba (stride=2)} & \textbf{Unified} & \textbf{\checkmark} & \textbf{\checkmark} & \textbf{63.4} \\ 
            
            \bottomrule
        \end{tabular}%
    }
\end{table}

\subsection{Ablation Study}

We conduct a comprehensive ablation study on LAV-DF to validate our 
core components and, critically, the architectural synergy between 
unified sequence fusion and the Mamba backbone. Results are reported 
as AP@0.95 in Table~\ref{tab:ablation_full}.

\subsubsection{Backbone-Fusion Compatibility}

\textbf{A Critical Negative Result.}
To validate that unified sequence representation is not universally 
beneficial but requires architectural compatibility, we conducted a 
controlled experiment: applying unified sequence fusion to the 
Transformer-based UMMAFormer~\cite{zhang2023ummaformer} baseline.

\textbf{Results:}
As shown in the first block of Table~\ref{tab:ablation_full}, 
UMMAFormer with unified sequence achieves only 34.6\% AP@0.95 
(with modality embeddings), \textit{underperforming} its original 
channel-concatenation baseline (37.6\%, $-3.0\%$ absolute). In stark 
contrast, our Mamba-based framework with the same unified fusion 
achieves 63.0\% ($+82\%$ relative improvement over Transformer+Unified).

\textbf{Analysis:}
This counterintuitive result reveals a fundamental insight: 
\textit{unified sequence representation demands architectural 
compatibility}. We identify three factors explaining Transformer's failure:

\begin{enumerate}
    \item \textbf{Quadratic Complexity Bottleneck:} Doubling sequence length from $T$ to $2T$ increases self-attention complexity from $O(T^2)$ to $O(4T^2)$, causing training instability and underfitting on long sequences.
    
    \item \textbf{Position Encoding Mismatch:} Transformer's sinusoidal position encoding assumes sequential temporal ordering. In unified sequences, audio frame at position $T$ and video frame at position $0$ represent the \textit{same} timestamp but are encoded as $T$ steps apart, breaking temporal correspondence.
    
    \item \textbf{Locality Bias:} Transformers learn locality bias during training. Unified sequences require modeling correlations between $\mathbf{v}_t$ (position $t$) and $\mathbf{a}_t$ (position $T+t$), spanning $T$ positions—beyond typical attention patterns.
\end{enumerate}

\textbf{Why Mamba Succeeds:}
In contrast, Mamba's state-space formulation naturally accommodates 
unified sequences through:
\begin{enumerate}
    \item \textbf{Linear complexity} $O(2T)$ scaling gracefully; 
    
    \item \textbf{Position-agnostic} state propagation $\mathbf{h}_t = \bar{\mathbf{A}}\mathbf{h}_{t-1}+\bar{\mathbf{B}}\mathbf{x}_t$ without absolute position dependence;
    
    \item \textbf{Selective scanning} with input-dependent gating adaptively focusing on intra-modal vs. cross-modal patterns.
\end{enumerate}
This ablation validates that our contributions are \textit{synergistic}: unified sequence fusion and Mamba backbone are mutually enabling, not independent improvements.

\subsubsection{Component-wise Analysis}

\textbf{Unified Sequence vs. Channel Concatenation.}
Comparing ActionMamba (39.1\%) with Variant C (43.3\%), switching to 
unified sequence yields $+4.2\%$ gain. This confirms that serializing audio-visual features enables better long-range cross modal dependency modeling by breaking rigid frame alignment constraints.

\textbf{Impact of Modality Embedding.}
Adding modality embeddings to unified sequence (UniSkip-Mamba full-scan) dramatically boosts performance to 63.0\% ($+19.7\%$ over Variant C). In contrast, adding them to channel concatenation (Variant B) degrades performance to 34.7\%. This indicates embeddings act as a critical compass for the unified sequence to distinguish modalities, whereas they introduce confusion in channel-separated inputs.

\textbf{Skip Scanning as Regularization.}
Applying stride-2 skip-scanning to our full model further improves 
AP@0.95 to 63.4\% ($+0.4\%$). However, applying it to channel 
concatenation (Variant A) hurts performance (36.7\%, $-2.4\%$ vs. 
ActionMamba). This suggests skip-scanning works synergistically with 
unified sequences as a temporal regularizer, filtering high-frequency noise while preserving context, but causes information loss in rigid channel structures.

\subsection{Frequency Domain Analysis}
\label{ssec:freq_analysis}

To understand \textit{why} skip-scanning achieves superior robustness, we analyze which frequency bands contain discriminative information and how scanning stride influences frequency sensitivity on 1,000 AV-Deepfake1M samples.

\textbf{Experimental Setup.} We apply band-pass filters to isolate three frequency ranges in normalized frequency units: \textbf{Low} (0-0.05, semantic patterns like speaker identity), \textbf{Mid} (0.05-0.15, dynamic transitions like audio-visual synchronization), and \textbf{High} ($>0.15$, frame-level details like compression artifacts). We measure performance drop (mAP@0.95) when each band is removed during inference. Detailed methodology in \textbf{Supplementary}.

\begin{table}[t]
\centering
\caption{Frequency Band Ablation Study. Performance change when removing each frequency band. Negative values indicate performance drop; positive values indicate improvement.}
\label{tab:freq_ablation}
\setlength{\tabcolsep}{8pt}
\begin{tabular}{lccc}
\toprule
\textbf{Model} & \textbf{Remove Low} & \textbf{Remove Mid} & \textbf{Remove High} \\
& (0-0.05) & (0.05-0.15) & ($>0.15$) \\
\midrule
Stride 1 & -81.3\% & -92.8\% & -20.3\% \\
Stride 2 & -56.3\% & -72.4\% & -0.7\% \\
Stride 4 & -72.4\% & -79.7\% & \textbf{+1.4\%} \\
\bottomrule
\end{tabular}
\end{table}

\textbf{Key Findings.} Table~\ref{tab:freq_ablation} reveals three critical insights:

\textbf{(1) Low/Mid Frequencies are Critical.} Removing low or mid-frequency components causes catastrophic degradation (56-93\% drop), confirming that semantic-level patterns (scene context, speaker identity) and dynamic transitions (audio-visual synchronization, prosodic rhythms) are fundamental to forgery detection. Stride 1's extreme sensitivity to mid-frequency removal (-92.8\%) indicates heavy reliance on frame-to-frame dynamics, which can be both a strength (capturing subtle transitions) and a weakness (vulnerable to noise).

\textbf{(2) High Frequencies Introduce Noise.} High-frequency removal has minimal impact (-0.7\% to +1.4\%). Remarkably, stride 4 \textit{improves} by \textbf{+1.4\%} when high frequencies are removed, providing strong evidence that fine-grained frame-level details contribute little to discrimination and may even introduce noise that hinders performance. This validates skip-scanning as beneficial regularization rather than information loss.

\textbf{(3) Nyquist Criterion Validation.} The results align perfectly with sampling theory. Stride=4 (Nyquist frequency $f_N=0.125$ Hz) shows improvement when high frequencies removed but suffers -79.7\% drop when mid frequencies removed—precisely because $f_N=0.125 < 0.15$ (discriminative band upper bound), causing aliasing and information loss. In contrast, stride=2 ($f_N=0.25$ Hz) shows only -0.7\% change for high-frequency removal and -72.4\% for mid-frequency removal—the smallest degradation among all models. This confirms stride=2 provides the \textit{optimal sampling rate}: fully preserving the discriminative band (0-0.15 Hz) while naturally attenuating high-frequency noise.

\textbf{Design Implications.} These findings provide theoretical justification for our skip-scanning design from signal processing perspective. The optimal stride ($p=2$) is not an arbitrary hyperparameter but a principled decision derived from the Nyquist-Shannon sampling theorem applied to the empirically-determined discriminative frequency band. This explains why stride=2 simultaneously improves accuracy, robustness, and efficiency—a frequency-aware design validated by both theory and experiments. Complete frequency spectrum analysis, noise sensitivity results, and cross-dataset comparisons are in \textbf{Supplementary}.

\section{Conclusion}

In this paper, we presented \textbf{UniSkip-Mamba}, a novel and efficient framework for AV-TFL. By rethinking the temporal modeling paradigm from a frequency-domain perspective, we addressed the twin challenges of computational efficiency and robust cross-modal learning. Our proposed \textbf{Unified Sequence Representation} breaks the rigid frame-alignment constraints of traditional channel concatenation, enabling flexible modeling of asynchronous cross-modal dependencies. Furthermore, our \textbf{Efficient Skip-Scanning Mamba Block} introduces a configurable stride mechanism that not only accelerates inference by up to 6$\times$ but also acts as a structural regularizer, naturally filtering out high-frequency noise while preserving discriminative forgery patterns. Extensive experiments on LAV-DF and AV-Deepfake1M benchmarks demonstrate that UniSkip-Mamba achieves state-of-the-art localization accuracy, particularly in high-precision regimes, while significantly reducing computational costs.

\textbf{Limitation and Future Work.} While our skip-scanning strategy excels at capturing long-range dependencies and suppressing noise, it may inherently attenuate extremely short-duration artifacts (e.g., single-frame glitches) that reside purely in the high-frequency band. Future work could explore adaptive stride selection mechanisms or multi-scale fusion strategies to dynamically balance fine-grained detection with long-term context modeling. Additionally, extending this unified efficient framework to other multi-modal understanding tasks, such as audio-visual event localization, remains a promising direction.

\vfill
\bibliographystyle{IEEEtran}   
\bibliography{references}      %

\end{document}